\documentclass[letterpaper, 10 pt, conference]{ieeeconf} 
\makeatletter
\let\NAT@parse\undefined
\makeatother

\makeatletter
\renewcommand{\paragraph}[1]{%
  \par\vspace{0.4em}%
  \noindent\textbf{#1.}%
}
\makeatother                                                      

\IEEEoverridecommandlockouts                              

\title{\LARGE \bf
Preparation of Papers for IEEE CSS Sponsored Conferences \& Symposia
}

\title{\LARGE \bf Accelerating Nonlinear MPC by Behavior Cloning with Separated Timescales:
Application to Nuclear Reactor Load-Following%
\thanks{A preliminary short version of this work has been accepted and will be presented at 
the NeurIPS 2026 Workshop ``AI Foundations for Power Grids: From Models to Deployment at Scale'' 
(non-archival), \url{https://ai4powergrids.github.io/cfp.html}.
The present paper substantially extends that work with
20 random seeds, in-distribution and out-of-distribution experiments, and a safety-augmented 
pipeline~\cite{hose2025approximate}.}}
\usepackage[utf8]{inputenc}
\usepackage[T1]{fontenc}      
\usepackage{graphicx}      
\usepackage{amsmath}               
\allowdisplaybreaks
\usepackage{amssymb}               
\usepackage{amsfonts}              
\usepackage{amsbsy}
\usepackage{bbm}
\usepackage{tikz}                  
\usepackage{multicol}
\usepackage{graphicx}
\usepackage{threeparttable}
\usepackage{soul,derivative,stmaryrd,mathtools,enumerate,enumitem,mathtools}
\usepackage{comment}
\usepackage{algorithm}
\usepackage{algorithmicx}
\usepackage{algpseudocode}
\usepackage[colorlinks=true,citecolor=blue,urlcolor=blue]{hyperref}
\usepackage{xfrac}

\usetikzlibrary{arrows,positioning,shapes}
\usetikzlibrary{shapes.multipart}
\usetikzlibrary{arrows.meta}
\usetikzlibrary{decorations.pathreplacing}
\usetikzlibrary{arrows,intersections}
\usetikzlibrary{shapes,arrows}
\usetikzlibrary{fit,positioning}

\newcommand{\mbb}[1]{\mathbb{#1}}

\newcommand{\mrm}[1]{\mathrm{#1}}

\newcommand{\R}{\mbb{R}}

\newcommand{\AO}{\mrm{AO}}
\newcommand{\I}{\mrm{I}}
\renewcommand{\P}{\mrm{P}}

\newcommand{\Xe}{\mrm{X}}

\renewcommand{\d}{\mrm{d}}
\newcommand{\Cbrcv}{\mrm{C}_\mrm{b}^{\mrm{rcv}}}
\newcommand{\Cbrcp}{\mrm{C}_\mrm{b}^{\mrm{rcp}}}
\newcommand{\Hbf}{\mrm{H}_\mrm{in}}
\newcommand{\Xbank}{\mrm{X}_\mrm{bank}}

\DeclareMathOperator*{\argmin}{arg\,min}

\DeclareMathSymbol{\shortminus}{\mathbin}{AMSa}{"39}

\usepackage{subcaption} 
\usepackage[table]{xcolor}

\definecolor{bestgreen}{RGB}{198, 230, 201} 
\definecolor{secondblue}{RGB}{187, 222, 251} 
\definecolor{lightgray}{RGB}{210, 210, 210} 

\definecolor{couleur1}{HTML}{0f4c5c}
\definecolor{couleur2}{HTML}{fb8b24}
\definecolor{couleur3}{HTML}{5f0f40}

\definecolor{yellow}{rgb}{1,1,0}
\definecolor{lightgreen}{rgb}{0.5,1,0.5}
\definecolor{darkgray}{rgb}{0.15,0.15,0.15}   
\definecolor{lightgray}{rgb}{0.94,0.94,0.94}  
\definecolor{lightlightgray}{rgb}{0.97,0.97,0.97}  
\definecolor{lightlightblue}{rgb}{0.95,0.97,1.00}  
\definecolor{darkred}{rgb}{0.80,0.00,0.00}    
\definecolor{darkgreen}{rgb}{0.00,0.70,0.00}    
\definecolor{darkblue}{rgb}{0.00,0.00,0.70}    
\definecolor{violet}{rgb}{0.35,0.00,0.70}

\definecolor{lightblue}{cmyk}{0.25,0.15,0,0}
\definecolor{lightred}{cmyk}{0,0.25,0.15,0}
\definecolor{lightgreen}{cmyk}{0.15,0,0.25,0}
\definecolor{lightyellow}{cmyk}{0,0,0.25,0}

\definecolor{darkyellow}{rgb}{0.9,0.9,0.00}    
\definecolor{orange}{rgb}{1.00,0.50,0.00}    
\usepackage{booktabs}

\author{%
 Perceval Beja-Battais \quad
 \thanks{P. Beja-Battais and N. Vayatis are with Université Paris-Saclay, ENS Paris-Saclay, CNRS, Centre Borelli, 91190, Gif-sur-Yvette, France 
        {\tt\small name.surname@ens-paris-saclay.fr}}
Alain Grossetête 
\quad
\thanks{A. Grossetête is with Framatome, Reactor Process Department, 92400, Courbevoie, France 
        {\tt\small alain.grossetete@framatome.com}}
Nicolas Vayatis
}

\begin{document}

\maketitle
\thispagestyle{empty}
\pagestyle{empty}

\begin{abstract}
Behavior Cloning (BC) of expert controllers for multiscale dynamical systems is challenging: state variables evolving at widely separated timescales induce highly nonlinear couplings that a shared latent space may fail to adequately represent, degrading closed-loop generalization. We propose Structured Representation Learning for Behavior Cloning (SRLBC), an architecture designed to leverage the physical knowledge of multiscale systems in which slow state variables, fast state variables, and predictable exogenous inputs evolve at different timescales. On 500 long-horizon roll-outs of a pressurized water reactor (PWR) load-following benchmark — a high-dimensional, stiff industrial system with challenging safety constraints — and across 20 training seeds, separated embeddings systematically outperform shared embedding baselines on cumulative cost, imitation accuracy and constraint satisfaction. The resulting standalone policies approach NMPC closed-loop performance at a computational cost 15 times lower on average. Since standalone policies still leave a few percent of trajectories infeasible, we embed the learned policies in a safety-augmented approximate NMPC framework, in which the optimizer is only called when the predicted input sequence is not certified as safe. This yields controllers still 6.8 times faster than the NMPC controller on average, with a feasibility (i.e., constraint satisfaction) of at least $99.8\%$ for all architectures, SRLBC retaining the lowest closed-loop cost. Under distribution shift, standalone BC policies degrade sharply: SRLBC degrades the least, and once safety-augmented, it retains a feasibility above $98\%$ with the lowest closed-loop cost. Beyond the quantitative results, Jacobian-based sensitivity analysis shows that the resulting controller relies on a sparse and less redundant set of latent features, each attributable to a single group of physical variables, and aligned with the physical timescale decomposition of the system.
\end{abstract}

\section{Introduction}

For the past decades, sophisticated controllers have been designed to meet industrial requirements. 
Model Predictive Control (MPC) is a widely adopted methodology~\cite{qin2003survey} that repeatedly solves an Optimal Control Problem (OCP) 
minimizing a cost function subject to dynamics and constraints. 
Nonlinear MPC (NMPC) extends this to nonlinear settings, but at the cost of significant online computational burden~\cite{rao2009survey, diehl2009efficient}, 
which limits its practical deployment in resource-constrained industrial pipelines~\cite{fiedler2023mpc}.

\paragraph{ML for MPC} Machine Learning (ML) has emerged as a principled means of reducing this online cost, 
giving rise to the Learning-Based MPC literature~\cite{hewing2020learning}. 
Existing approaches either learn surrogate models of the dynamics~\cite{hossain2023machine, beja2026accelerating} or directly approximate the 
MPC controller~\cite{vaupel2020accelerating, li2024faster}.
Behavior Cloning (BC) is a simple methodology consisting in mimicking the behavior of an expert through supervised learning.
Such methods offer real-time deployment without online optimization, while retaining near-optimal performance when the learned policy generalizes well~\cite{wu2025tutorial}.
Safety-augmented neural networks~\cite{hose2025approximate} combine the best of both worlds: 
the output of a neural network policy, e.g., trained through BC, is applied to the system whenever it is certified as safe, 
and the MPC expert is called otherwise.

\begin{figure}
\centering
\resizebox{0.9\columnwidth}{!}{%
\begin{tikzpicture}[auto, node distance=2.7cm, >=latex']

    \tikzstyle{block} = [draw, rectangle, minimum height=1.2cm, minimum width=2cm]
    \tikzstyle{block_vert} = [draw, rectangle, minimum height=1cm, minimum width=0.4cm]
    \tikzstyle{sum} = [draw, circle, inner sep=1pt]
    
    \node[circle, draw=black, fill=darkblue!30] (x) at (0,1.6) {$x$};
    \node[circle, draw=black, fill=darkgreen!30] (z) at (0,0) {$z$};
    \node[circle, draw=black, fill=darkred!30] (w) at (0,-1.6) {$w$};
    
    \node[draw, rectangle, right of=x, fill=lightblue!30, align=center, minimum height=1.2cm, minimum width=3.7cm] (B1) {Slow State Encoder\\$\phi_x$};
    \node[right of=B1, node distance=3cm, label=$\phi_x(x)$] (z_x) {};

    \node[draw, rectangle, right of=z, fill=lightgreen!30, align=center, minimum height=1.2cm, minimum width=3.7cm] (B2) {Fast State Encoder\\$\phi_z$};
    \node[right of=B2, node distance=3cm, label=$\phi_z(z)$] (z_z) {};

    \node[draw, rectangle, right of=w, fill=lightred!30, align=center, minimum height=1.2cm, minimum width=3.7cm] (B3) {Exogenous State Encoder\\$\phi_w$};
    \node[right of=B3, node distance=3cm, label=$\phi_w(w)$] (z_w) {};

    \node[block_vert, right of=z_z, fill=gray!30, align=center] (concat) {Latent Representation\\$\Phi(x, z, w)$};
    
    \node[block, right of=concat, node distance=3.5cm, fill=darkyellow!30, align=center] (B4) {Controller\\$\pi$};
    \node[right of=B4, node distance=2.2cm, inner sep=0pt, minimum size=0pt, label={$\pi\circ\Phi(x, z, w)$}] (u) {};
    
    \node[draw=none, inner sep=0pt, minimum size=0pt] (block_before) at (-1,-1.6) {};

    \node[block, fill=gray!30] (env) at (6,-3) {Environment};

    \draw[->] (x) -- (B1.west);
    \draw[->] (z) -- (B2.west);
    \draw[->] (w) -- (B3.west);
    
    \draw[->] (B1) -| (concat.north);
    \draw[->] (B2) -- (concat.west);
    \draw[->] (B3) -| (concat.south);
    \draw[->] (concat.east) -- (B4.west);
    \draw[-] (B4) -- (u);

    \draw[->] (u) -- +(0, -3) -- (env);
    \draw[-] (env) -- +(-7, 0) -- (block_before);
    \draw[->] (block_before) -- +(0, 3.2) -- (x);
    \draw[->] (block_before) -- +(0, 1.6) -- (z);
    \draw[->] (block_before) -- (w);

\end{tikzpicture}%
}
\caption{Proposed multi-stream architecture: we encode the input data in separated latent spaces 
and the controller $\pi$ acts on the product latent space.}
\label{fig:summarize_controller}
\end{figure}
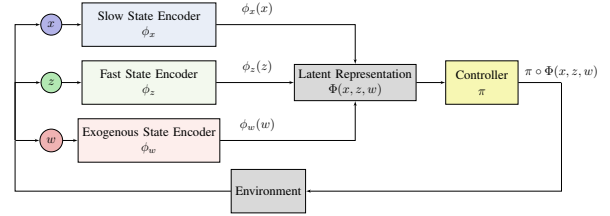

\paragraph{Representation Learning for Behavior Cloning} For BC, a critical ingredient for generalization is the quality of the state representation fed to the controller. 
Representation learning~\cite{bengio2013representation} provides tools to embed state variables into
task-relevant representations~\cite{lesort2018state}. 
However, what constitutes a good representation depends on multiple, sometimes competing criteria: expressivity, preservation of geometric structure, 
task performance, and interpretability~\cite{charte2020analysis}.
Prior works, such as Embed to Control (E2C)~\cite{watter2015embed}, leverage representation learning for control systems 
while imposing linear dynamics over a shared latent space. 
This constraint can be both restrictive and ill-suited to complex industrial systems.
This limitation is most severe for multiscale dynamical systems, which arise naturally in physics and engineering~\cite{horstemeyer2009multiscale}. 
When state variables evolve at widely separated timescales, a single shared embedding may conflate slow and fast dynamics. 
This can induce a degraded representation and a less interpretable controller.
In control theory, exploiting timescale separation in the controller structure is well established: 
composite control~\cite{kokotovic1999singular} and MPC schemes for singularly perturbed systems~\cite{chen2012composite, ellis2013economic} 
decompose the design accordingly. 
However, these methods still optimize online. 
In this work, we transpose this principle to the learned representation of an explicit policy.
In practice, multiscale systems are typically stiff, preventing naive numerical integration. 
They can also exhibit nonsmooth, nonconvex cost landscapes that challenge standard solvers~\cite{mao2018successive, nurkanovic2023numerical}. 
For neural networks, structuring a latent representation into distinct components is not new in itself: 
disentangled representation learning recovers statistically independent 
generative factors in an unsupervised manner~\cite{locatello2019challenging}, 
and multi-modal architectures separate encoders across heterogeneous input modalities~\cite{radford2021learning, shafiullah2022behavior}. 
In contrast, our decomposition is neither discovered in a task-agnostic manner nor imposed across data modalities: 
it is prescribed by the known timescale separation of the physical system, 
a structural property already available from prior modeling. 
The performance of this decomposition is evaluated directly on closed-loop control performance and fidelity to an expert controller, 
rather than on reconstruction or disentanglement metrics.

\paragraph{Contributions} To this end, we introduce Structured Representation Learning for 
Behavior Cloning (SRLBC), a simple architecture for BC of expert controllers 
of multiscale nonlinear dynamical systems, where state variables evolve at widely separated timescales. 
Our contributions are as follows.

\paragraph{A timescale-aware multi-stream architecture} 
We encode slow state variables, fast state variables, 
and exogenous input variables into separate latent spaces. 
Then, a structured closed-loop controller is trained over the product latent space.
The architecture is summarized in Fig.~\ref{fig:summarize_controller}.
Separating timescales improves closed-loop performance compared to a shared embedding baseline, 
and yields a controller relying on fewer, less redundant latent features. 
These gains are significant and consistent across 20 training seeds.

\paragraph{Safe deployment via safety augmentation}
Embedding the learned policies within the safety-augmented approximate NMPC framework of~\cite{hose2025approximate} 
calls the optimizer at less than 1.6\% of the timesteps, yielding a 6.8-fold speedup over the NMPC controller with a feasibility of at least $99.8\%$. 
In this configuration, SRLBC also achieves a lower closed-loop cost than the baselines, including on out-of-distribution transients.

\paragraph{Validation on an industrial benchmark} 
We demonstrate the efficiency of SRLBC on a challenging full-scale industrial benchmark: 
optimal load-following control of a PWR.
We also conduct an interpretability analysis: separated embeddings yield sparser, more interpretable 
and less redundant representations than shared ones, aligned with the system's physics (Fig.~\ref{fig:sensitivity_analysis}).

This work is organized as follows. 
Section~\ref{sec:PbStatement} defines the problem structure and reviews related work such as approximate MPC, 
latent-space control, and imitation learning. 
Section~\ref{sec:Model} presents the architecture and training procedure. 
Section~\ref{sec:Experiment} details the experimental validation. 
Section~\ref{sec:Discussion} discusses results, limitations, and perspectives.

\section{Problem Statement \& Related Work}\label{sec:PbStatement}

\paragraph{Multiscale control problem}
We consider multiscale systems~\cite{kuehn2015multiple} subject to an exogenous signal $w(t)$ known in advance, 
which yields the general fast-slow system
\begin{equation}\label{eq:full_dynamics}
    \left\{
    \begin{aligned}
        &\dot{x}(t) = f(x(t), z(t), w(t), u(t)), \quad &&x(0) = x_0, \\
        &\varepsilon\dot{z}(t) = g(x(t), z(t), w(t), u(t)), \quad &&z(0) = z_0,
    \end{aligned}
    \right.
\end{equation}
where the state $[x(t)^\top, z(t)^\top]^\top$ decomposes into slow and fast components, 
with timescales separated by the stiffness parameter $0 < \varepsilon \ll 1$.
\newcommand{\MPC}{{\scriptscriptstyle\mathrm{MPC}}}
\newcommand{\BC}{{\scriptscriptstyle\mathrm{BC}}}

Our aim is to design an explicit state-feedback controller, without online optimization,
mimicking the behavior of an expert control input $u^*(t)$, obtained by solving the nonlinear OCP
\begin{equation}\label{eq:OCP}
    \left\{
    \begin{aligned}
        &u^*\!\coloneqq\!\argmin_{u}\!\left[J(u)\!\coloneqq\!\int_{t_0}^{t_f}\!\ell(x(t), z(t), w(t), u(t)) \d t\right], \\
        &\text{s.t.~\eqref{eq:full_dynamics} and $h(x(t), z(t), w(t)) \leq 0$.}
    \end{aligned}
    \right.
\end{equation}
In the following, we assume that~\eqref{eq:OCP} can be solved by an NMPC scheme, 
which defines the NMPC controller $\pi_\MPC(x_0, z_0, w)$.
However, this scheme may be too slow for embedded deployment in industrial processes, 
which is why we aim to imitate it.

\paragraph{Representations for control}
An encoder $\Phi$ is a learned transformation of input variables
revealing the information relevant to a specified task~\cite{lesort2018state}. 
E2C~\cite{watter2015embed} learns representations of the state variables in which the dynamics are approximately linear, 
enabling standard MPC tools. 
We choose instead to prioritize the preservation of the multiscale structure: 
we encode the slow, fast and exogenous variables with three separate encoders. 
We consider latent representations of sufficiently large dimension (but not unnecessarily large).
Indeed, an overly restrictive bottleneck could discard complex couplings pertinent for designing control actions, 
while a larger representation does not necessarily improve control performance (see Section~\ref{sec:Experiment}, 
larger baseline). 


\paragraph{BC and safety} 
Imitation Learning (IL) consists of a class of methods in which a policy $\pi$ is learned by observing an expert's demonstration~\cite{hua2021learning}. 
The simplest IL framework is BC~\cite{torabi2018behavioral}.
It assumes access to a dataset of expert control inputs and trajectories $\mathcal{D} \coloneqq \{(x_i, z_i, w_i, \pi_{\MPC}(x_i, z_i, w_i)), i\in \llbracket 1, N\rrbracket\}$, 
and learns a mapping $\pi_\BC \approx \pi_\MPC$.
However, learning such policies for complex systems can be challenging: a standard issue is the offline computational cost required to generate a representative 
set of demonstrations from $\pi_\MPC(x,z,w)$. 
To safely deploy a learned clone of $\pi_\MPC$, one can either (i) define a Robust MPC (RMPC) problem that ensures statistical (safety) guarantees on the clone~\cite{hertneck2018learning},
or (ii) integrate the policy inside a safety pipeline calling the NMPC controller if the neural policy is not safe (e.g., when it does not satisfy the constraints)~\cite{hose2025approximate}.
In some industrial contexts, (i) may be out of reach: 
the assumptions required by statistical learning theory can be hard to verify, or the resulting bounds too loose to yield applicable guarantees. 
Moreover, for safety-critical applications, we emphasize the interpretability of $\pi_\BC$, at the expense of 
the network's complexity --- and thus of its statistical guarantees.

\section{Proposed Framework} \label{sec:Model}

\paragraph{Designing and training the controller}
Our model consists of four components:
a slow-state encoder $\phi_x : \mathbb{R}^{n_x} \rightarrow \mathbb{R}^{d_x}$,
a fast-state encoder $\phi_z : \mathbb{R}^{n_z} \rightarrow \mathbb{R}^{d_z}$,
an exogenous-variable encoder $\phi_w : \mathbb{R}^{n_w} \rightarrow \mathbb{R}^{d_w}$,
and a latent controller
$\pi_{\BC} : \mathbb{R}^{d_x+d_z+d_w} \rightarrow \mathbb{R}^{n_u}$.
The entire architecture, summarized in Fig.~\ref{fig:summarize_controller}, is trained via BC to approximate the NMPC controller
$\pi_{\MPC}(x,z,w)$ by minimizing the BC loss
\begin{equation}\label{eq:BC_loss}
    \ell_{\BC}(\cdot)
    \coloneqq
    \left\|
    \pi_{\BC}\circ\Phi(\cdot)
    -
    \pi_{\MPC}(\cdot)
    \right\|,
\end{equation}
where $\Phi \coloneqq [\phi_x^\top, \phi_z^\top, \phi_w^\top]^\top$.

Compared with a naive shared embedding, this architectural prior constrains the encoding, not the decision: 
each encoder summarizes a single group of variables, while the nonlinear controller $\pi_\BC$ 
can still represent arbitrary couplings between the latent blocks. 
The prior is thus that couplings should be resolved at the controller level, 
rather than mixed from the first layer. 
Although a sufficiently expressive shared embedding could, in principle, recover the same decomposition, 
explicitly enforcing it substantially reduces the space of admissible representations. 
We motivate this prior for our application in Section~\ref{sec:Experiment}.

\paragraph{Enforcing decorrelation}
We do not restrict the controller to be linear as in E2C~\cite{watter2015embed}.
Still, it should be designed to be as simple and expressive as possible for the sake of interpretability.
To this end, enforcing independence between the components of the embedding can be relevant.
This can be done in practice during training by learning decorrelated components in the embeddings, 
e.g., by minimizing $\|\mathrm{Cov}(\phi_i) - I_{d_i}\|, \ i \in \{x, z, w\}$.
We investigate this hypothesis on our structured encoder.

\paragraph{Finetuning with DAgger}
Dataset Aggregation (DAgger)~\cite{ross2011reduction} is an iterative algorithm for policy refinement that collects the expert's 
actions along trajectories generated by the current policy, thereby accumulating new data points in regions where it may make errors. 
One round of DAgger is included in the training pipeline of every policy considered in this work, baselines included.

\definecolor{oursblue}{RGB}{230,240,250}
\definecolor{bestgreen}{RGB}{220,245,220}
\definecolor{secondyellow}{RGB}{255,245,210}

\section{Experiments}\label{sec:Experiment}

\begin{figure*}
    \centering 
    \includegraphics[width=0.74\textwidth]{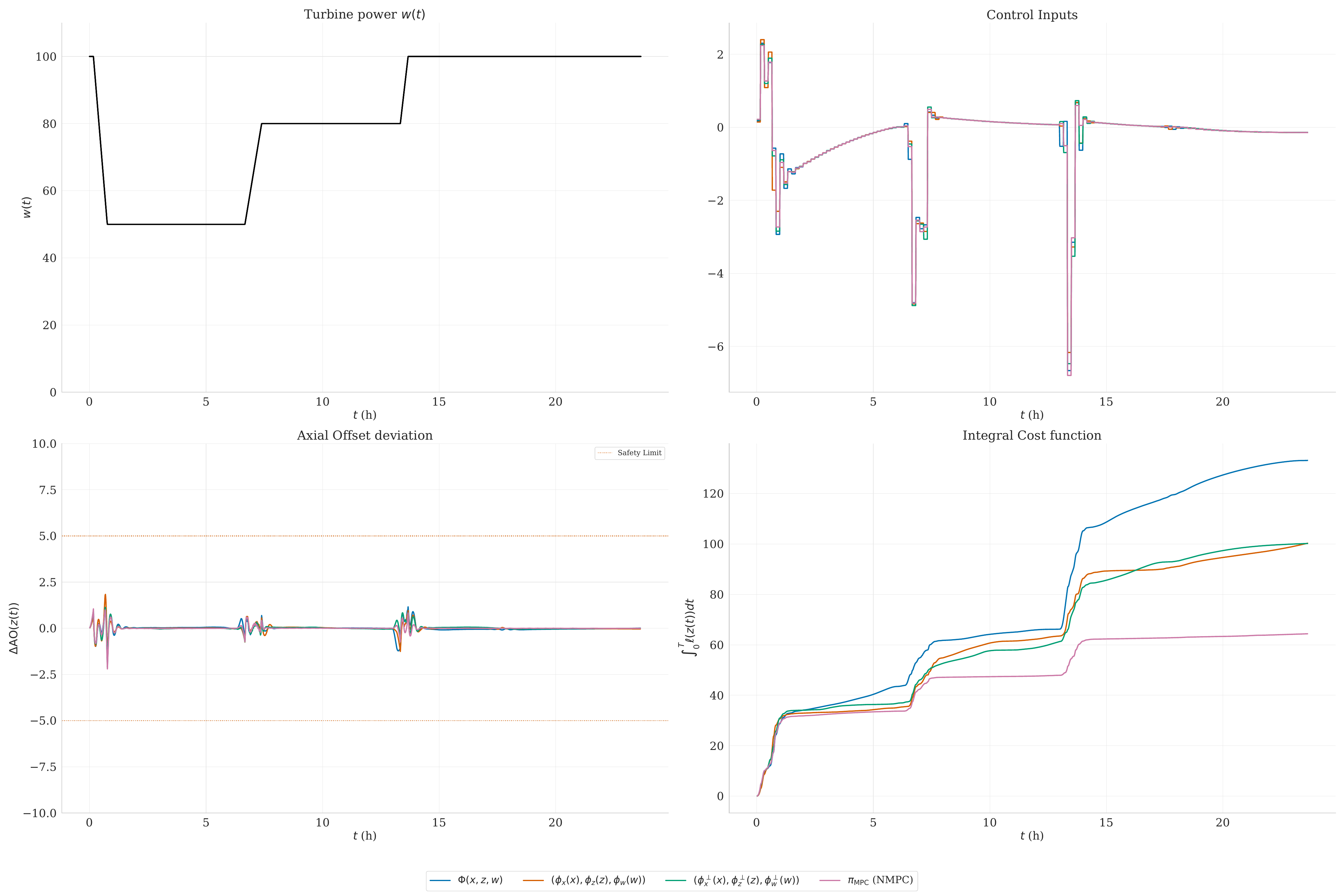}
    \caption{Results of the NMPC controller (pink)
    vs. several methodologies on a 24h example transient. 
    In blue, we plot the controller trained on a shared embedding (BC Baseline), in orange,
    trained on the structured embedding (SRLBC) and in green on structured embeddings with enforced decorrelation (SRLBC + Decor.).
    Top-left: turbine power, i.e., exogenous input variable $w(t)$. Top-right: control input sequences of each method.
    Bottom-left: axial offset deviation, linked to both the stage cost function
    $\ell(z(t), w(t))$ and safety limit $h(z(t), w(t)) \leq 0$. Bottom-right: integral cost function $\int_{0}^{t} \ell(z(s), w(s))\,\d s$.}
    \label{fig:example}
\end{figure*}

\paragraph{System dynamics}
The goal is to design a closed-loop controller for the load-following of PWRs.
We use a simplified version of the model from~\cite{dupre2021enhanced}.
The dynamical system is described by the slow state variable $x(t) = (\I(t), \Xe(t), \Cbrcv(t), \Cbrcp(t))$, gathering the iodine and xenon densities and the boron concentrations, 
the fast state variable $z(t) = (\Hbf(t), \Xbank(t), \P(t))$, gathering the inlet enthalpy, the control rod position and the neutronic power in each axial mesh, 
and the exogenous input variable $w(t) = P_\mathrm{turb}(t)$, i.e., the turbine power profile, known 30 minutes in advance.
Its dynamics are given by\footnote{
    $C_\mathrm{b, inj}$ follows the equation
            \begin{equation*}
                    C_\mathrm{b, inj}(t) \coloneqq \left\{\begin{aligned}
                    \Cbrcp(t) + \frac{u(t)}{Q_{\mathrm{ch}}}\left(C_{\mathrm{bor, rea}} - \Cbrcp(t)\right)~\text{if $u(t) \geq 0$,} \\
                    \Cbrcp(t) - \frac{u(t)}{Q_{\mathrm{ch}}}\left(C_{\mathrm{dil, rea}} - \Cbrcp(t)\right)~\text{if $u(t) < 0$,}
                \end{aligned}\right.
            \end{equation*}
                    }
    \footnote{$\mathrm{GF4}$ is a discontinuous profile. 
    All the other undefined parameters are either scalar or matrix constants.}:
\begin{subequations}\label{eq:dynmical_system}
    \begin{align}
        \frac{\d \I}{\d t}\!&=\!N\gamma_\I \P(t) - \lambda_\I \I(t), \\
        \frac{\d \Xe}{\d t}\!&=\!N\gamma_\Xe \P(t)\!+\!\lambda_\I \I(t)\!-\!(\lambda_\Xe\!+\!\sigma_\Xe N \P(t))\Xe(t), \\
        \frac{\d \Cbrcv}{\d t}\!&=\!\frac{1}{\tau_{\mathrm{rcv}}}\left(-\Cbrcv(t) + C_\mathrm{b, inj}(t)\right), \\
        \frac{\d \Cbrcp}{\d t}\!&=\!\frac{Q}{M_{\mathrm{rcp}}}(\Cbrcv(t) - \Cbrcp(t)), \\
        \frac{\d \Hbf}{\d t}\!&=\!\frac{K_\mathrm{H}}{\tau_{\mathrm{sg}}}\left(\sum_{i=1}^{N}\P_i(t) - P_\mathrm{turb}(t)\right), \\
        \frac{\d \Xbank}{\d t}\!&=\!\mathrm{GF4}(T_{\mathrm{ref}}(P_\mathrm{turb}(t)) - T(\Hbf(t))), \\
        \frac{\d \P}{\d t}\!&=\!\left(\frac{(\rho(t) - \beta) I_{N} + D_{\mathrm{exch}}}{\Lambda^*}\right) \P(t),
    \end{align}
\end{subequations}
where $\I, \Xe, \P \in \R^6$ are discretized along the core vertical axis. 
The control input variable, $u(t)$, represents dilutions or borations from a human operator. 
This system is highly nonlinear due to multiple factors: stiffness arises from 
the fast-evolving coupled subsystem $z = (\Hbf, \Xbank, \P)$, with challenging nonlinearities 
induced by the reactivity $\rho(t)$ (a highly nonlinear function, depending on all the state variables of the system~\cite{reuss2003precis, stacey2018nuclear}).
Typically, the fast dynamics from system~\eqref{eq:dynmical_system} reach equilibrium on timescales shorter than one second, 
whereas the slow dynamics evolve over timescales ranging from minutes to hours. 
Consequently, one can introduce a small stiffness parameter $\varepsilon \leq 10^{-2}$ representing 
the typical ratio between the two timescales.
Moreover, the control input variable $u(t)$ acts differently depending on its sign, 
and at $u(t) = 0$, the quantity $C_\mathrm{b, inj}(t)$ is not differentiable with respect to $u(t)$.
We use the notations $f, g$ from previous sections to designate the slow and fast parts of the dynamics, and 
refer to system~\eqref{eq:dynmical_system} for the specific equations.
The high dimension of the problem, with $n_x = 14, n_z = 8, n_w = 30$,
induces a non-trivial mapping $(x(t), z(t), w(t), \dots, w(t+k \ \d t)) \mapsto \pi_\MPC(x,z,w) \coloneqq (u^*(t), \dots, u^*(t+k \ \d t))$. 

\paragraph{Why separate the timescales in this system}
System~\eqref{eq:dynmical_system} is coupled but heavily structured. 
The boron subsystem $(\Cbrcv, \Cbrcp)$ is driven only by $u$, 
and $(\I, \Xe)$ act as low-pass filters of $\P$ with time constants of several hours: 
over the 30-min NMPC horizon, fast transients of $z$ barely influence $x$. 
Conversely, $z$ depends strongly on $x$ through $\rho$. 
Since $\varepsilon \ll 1$, $z$ quickly converges to a quasi-steady state $z^*(x, u, w)$~\cite{kokotovic1999singular}, 
and the remaining information it carries mostly describes fast transients. 
Encoding each timescale separately thus reflects the singularly perturbed structure of~\eqref{eq:full_dynamics}, 
while the coupling between the slow manifold and the fast transients is left to the controller.

\begin{figure*}[t]
    \centering
    \includegraphics[width=0.74\textwidth]{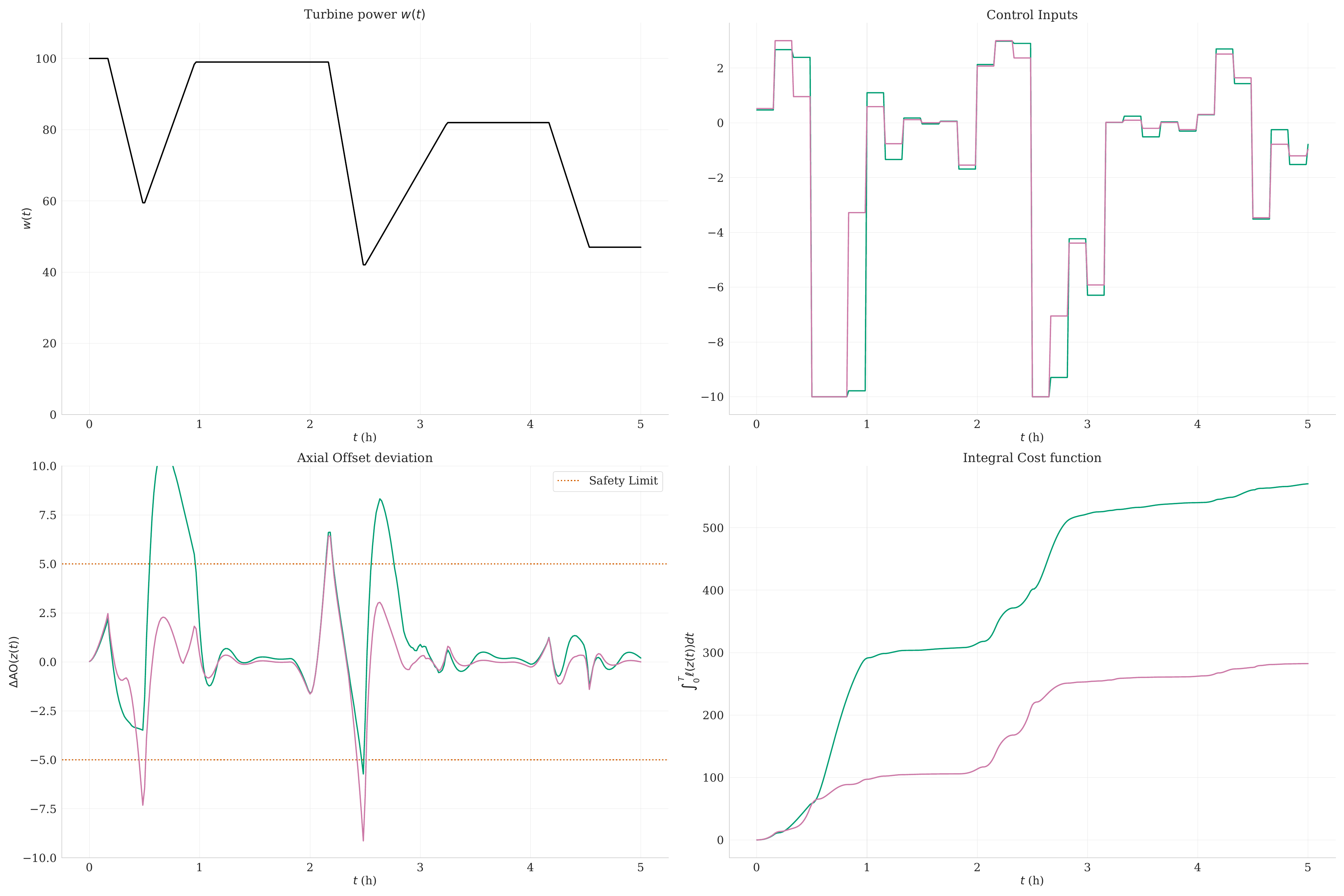}
    \caption{Out-of-distribution infeasible transient: the power setpoint abruptly changes direction during power variations (top-left). 
    Both the NMPC controller (pink) and the SRLBC (green) policies fail to satisfy the safety constraint throughout this transient.}
    \label{fig:ood}
\end{figure*}

\paragraph{Training data acquisition} \label{subsec:Data}
To design the dataset $\mathcal{D} \coloneqq \{(x_i, z_i, w_i, \pi_{\MPC}(x_i, z_i, w_i)),~i\in \llbracket 1, N\rrbracket\}$, 
a large number of simulations were collected in which the NMPC controller, in the spirit of~\cite{dupre2021enhanced}, solves~\eqref{eq:OCP} with 
\begin{equation*}
    \begin{aligned}
        \ell(z, w) &\coloneqq \left|\AO(\P) - \AO_\mathrm{ref}(w)\right|^2, \\
        h(z, w) &\coloneqq \left|\AO(\P) - \AO_\mathrm{ref}(w)\right| - \Delta\AO_{\max},
    \end{aligned}
\end{equation*}
where $\AO(\P) \coloneqq \big(\sum_{i=1}^{3} \P_i - \sum_{i=4}^{6} \P_i\big)/\sum_{i=1}^{6} \P_i$ is the axial offset 
of the core power distribution, and $\Delta\AO_{\max} = 5\%$.
In practice, $\ell$ and $h$ penalize the same quantity, but in a different way: 
$\ell$ enforces low distance on average, whereas $h$ avoids strong oscillations during 
load-follows (see Figs.~\ref{fig:example} and~\ref{fig:ood}).
This OCP involves a non-convex, nonlinear cost function and non-convex nonlinear constraints over 
the complex multiscale system~\eqref{eq:dynmical_system}, which explains why statistical bounds on a BC controller
are very hard to derive.
The simulations were sampled using a random transient generator which produces realistic random exogenous sequences of $P_\mathrm{turb}$ (see~\cite{bejabattais2026SRLBC}).

\paragraph{NMPC formulation and solver}
The OCP is transcribed by single shooting over a horizon of $K = 30$ steps of $\d t = 1$ min. 
The control input is piecewise constant over three intervals of $10$ min, 
so that $\pi_\MPC$ outputs $(u^*_1, u^*_2, u^*_3)$ with $u^*_j \in [-10, 3]$ kg/s~\cite{dupre2021enhanced}, 
and the constraint $h \leq 0$ is enforced at every step. 
The stiff dynamics are integrated with a dedicated stiff solver, and the resulting nonlinear program is solved 
with IPOPT~\cite{wachter2006implementation} via CasADi~\cite{andersson2019casadi} 
(tolerance $0.1$, at most $30$ iterations for fast inference). 
Depending on the state and exogenous variables, a single solve can take up to 1 min. 
In total, approximately $80\,000$ training points of the form 
$(x_i, z_i, w_i, \pi_{\MPC}(x_i, z_i, w_i))$ were collected through simulations of approximately $260$ trajectories of $48$ hours, 
for a cumulated offline cost of approximately $24$ hours ($5$ to $6$ minutes of wall-clock time per trajectory on average). 
The DAgger round then expands the training set of each policy to approximately $115\,000$ points (i.e., about 400 trajectories).

\begin{table*}[t]
\centering
\renewcommand{\arraystretch}{1.25}
\resizebox{\textwidth}{!}{%
{\small
\begin{tabular}{|c|cccc||cccc|}
\specialrule{1.2pt}{0pt}{0pt}
\textbf{Method}
& \multicolumn{4}{c||}{\textbf{Standalone (without Safety Augmentation)}}
& \multicolumn{4}{c|}{\textbf{Safety-Augmented~\cite{hose2025approximate}}}
\\
\specialrule{0.8pt}{0pt}{0pt}
\textbf{Embed. (Size)}
& \textbf{Cost Ratio to NMPC}
& \textbf{Rel. Gap to NMPC}
& \textbf{Constraint Satisfaction}
& \textbf{Time (s)}
& \textbf{Cost Ratio to NMPC}
& \textbf{Constraint Satisfaction}
& \textbf{Time (s)}
& \textbf{Rejection Rate}
\\
\specialrule{1.2pt}{0pt}{0pt}

BC Baseline (64)
& $1.166 \pm 0.029$ & $(9.9 \pm 1.0) \cdot 10^{-3}$ & $94.2 \pm 1.7\%$ & $20.2\pm 0.1$
& $1.094 \pm 0.022$ & $99.8 \pm 0.1\%$ & $46.3\pm 1.0$ & $1.59 \pm 0.03\%$
\\

BC Baseline (128)
& $1.130 \pm 0.017$ & $(8.8 \pm 0.4) \cdot 10^{-3}$ & $93.6 \pm 1.0\%$ & $20.2\pm 0.1$
& $1.074 \pm 0.010$ & $99.8 \pm 0.2\%$ & $46.2\pm 1.2$ & $1.56 \pm 0.02\%$
\\

\specialrule{0.8pt}{0pt}{0pt}

\rowcolor{oursblue}
SRLBC (64)
& $\mathbf{1.074 \pm 0.009}$ & $\mathbf{(7.0 \pm 0.4) \cdot 10^{-3}}$ & $\mathbf{97.4 \pm 0.7\%}$ & $20.3\pm 0.1$
& $\mathbf{1.053 \pm 0.005}$ & $\mathbf{99.9 \pm 0.1\%}$ & $\mathbf{45.6\pm 1.1}$ & $1.56 \pm 0.01\%$
\\

\rowcolor{oursblue}
SRLBC (64) + Decor.
& $1.083 \pm 0.012$ & $(7.5 \pm 0.4) \cdot 10^{-3}$ & $97.0 \pm 1.3\%$ & $20.3\pm 0.1$
& $1.057 \pm 0.006$ & $\mathbf{99.9 \pm 0.1\%}$ & $46.0\pm 1.2$ & $1.57 \pm 0.02\%$
\\

\specialrule{0.8pt}{0pt}{0pt}

NMPC
& $\mathit{1.00}$ & $\mathit{0}$ & $\mathit{100.0\%}$ & $\mathit{311.6 \pm 12.2}$
& $\mathit{1.00}$ & $\mathit{100.0\%}$ & $\mathit{311.6 \pm 12.2}$ & --
\\

\specialrule{1.2pt}{0pt}{0pt}
\end{tabular}
}}

\caption{Comparison between the different methodologies on 500 test roll-outs of 48h, on which the NMPC controller $\pi_\MPC$ is feasible by construction.
\textbf{Standalone:} the learned policy is applied directly. 
\textbf{Safety-augmented:} the policy output is applied when certified safe by forward simulation, and the NMPC controller is called otherwise~\cite{hose2025approximate}. 
The rejection rate is the proportion of timesteps at which the policy output is not certified.
Cost ratio to NMPC $=J(\pi_{\BC}\circ\Phi)/J(\pi_{\MPC})$, 
Rel. gap to NMPC $=\|\pi_{\BC}\circ\Phi-\pi_{\MPC}\|/\|\pi_{\MPC}\|$.
Constraint satisfaction (or feasibility) measures the proportion of 48h roll-outs respecting the constraint $h(z(t), w(t)) \leq 0$ 
at every timestep. 
Each metric is averaged over the test roll-outs, and results are displayed as $\mathrm{Mean} \pm \mathrm{Std}$ 
over 20 training seeds. 
For $\pi_\MPC$, the standard deviation of the computation time is taken over the 500 roll-outs.
Bold values indicate the best learned policy for each metric.
BC Baseline indicates the results with the shared encoder $\Phi(x, z, w)$, SRLBC with 
the proposed methodology $(\phi_x(x), \phi_z(z), \phi_w(w))$, and Decor. indicates 
that the embedding is enforced to learn decorrelated components.}

\label{tab:quantitative_results}
\end{table*}

\paragraph{Designing the encoders and the controller}
We used a 2-layer MLP with $d = 64$ neurons per hidden layer and ReLU activation for each encoder $\phi_x(x), \phi_z(z), \phi_w(w)$.
The controller block has the same architecture.
The encoders were pretrained on the reconstruction loss, after which the main training consists of 
simultaneously optimizing the encoders and controllers on the BC loss~\eqref{eq:BC_loss} in order 
to finetune the encoders specifically for control tasks.
The dimensions of the last embedding layer of the encoders were selected by comparing a validation BC loss over all 
possible 91 combinations in $\{(d_x, d_z, d_w), \ n_x \leq d_x, n_z \leq d_z, n_w \leq d_w, \ d = d_x + d_z + d_w\}$, 
so that the concatenated layer $[\phi_x(x)^\top, \phi_z(z)^\top, \phi_w(w)^\top]^\top$ contains representations 
of dimension larger than or equal to that of the input variables.
This latent-dimension search is specific to the separated architecture, 
which must allocate capacity across three constrained subspaces a priori. 
We compare the performance of the proposed methodology with a shared embedding of dimension $d$ on $(x, z, w)$.
For the shared baseline, no such decomposition needs to be chosen.
With this methodology, we found $d_x = 14, \ d_z = 12, \ d_w = 38$.
Shared embeddings and separated embeddings share the same total latent 
dimension $d = 64 = d_x + d_z + d_w$, ensuring a fair comparison.
We also compared our methodology against a larger encoder-controller baseline, with $2d = 128$ neurons per layer, 
and a $2d$-dimensional embedding.
Finally, we trained the same architecture on the hybrid loss given by 
\begin{equation*}
    \ell_\BC + \lambda \sum_{i\in \{x, z, w\}}\|\mathrm{Cov}(\phi_i) - I_{d_i}\|,
\end{equation*}
enforcing decorrelation, as stated previously (Decor. in Tables~\ref{tab:quantitative_results}--\ref{tab:infeasible}).
The hyperparameter $\lambda$ was fixed to $10^{-4}$ by optimizing over the log-grid $[10^{-5}, 10^{-1}]$. 

\paragraph{Evaluation datasets}
We evaluate the policies on two test sets. 
The in-distribution set contains new transients sampled from the same generator as the training set 
(Table~\ref{tab:quantitative_results}, and Table~\ref{tab:infeasible}, left). 
The out-of-distribution (OOD) set is produced by a modified transient generator, which draws abrupt changes in the setpoint $w(t)$ 
during power variations (e.g., Fig.~\ref{fig:ood}). 
The setpoint remains known 30 minutes in advance, but its direction can be reversed at short notice: 
this represents real-life scenarios in which an opposite directive had been planned and is changed within the last 30 minutes. 
Such transients are never seen during training and are harder to follow 
(Table~\ref{tab:ood}, and Table~\ref{tab:infeasible}, right).

\paragraph{In-distribution results}
Each architecture was trained with 20 different random seeds, and each resulting policy was evaluated 
on a test set of $500$ realistic feasible transients of 48h, i.e., load-following scenarios 
on which the NMPC controller satisfies the safety constraint and thus provides a valid reference.
Table~\ref{tab:quantitative_results} reports the mean and standard deviation across seeds. 
SRLBC reduces the excess closed-loop cost with respect to the NMPC controller from $16.6\%$ ($13.0\%$ for the larger baseline) to $7.4\%$, 
and the gap to its control inputs by $30\%$, further supporting the 
idea that separated embeddings help the learned policy to better approximate the NMPC controller. 
It also raises standalone feasibility from $94.2\%$ to $97.4\%$. 
The differences between SRLBC variants and both baselines in cost, imitation accuracy and feasibility are significant (one-sided Mann-Whitney tests across seeds with Holm correction, $p < 10^{-4}$), 
and SRLBC is also markedly less sensitive to the training seed (standard deviation of the cost ratio divided by 2 to 3). 
Enforcing decorrelation preserves most of these gains, at a marginal loss in cost and imitation accuracy 
and with a comparable feasibility ($97.0\%$ vs. $97.4\%$): 
its benefit lies in the structure of the representation rather than in closed-loop performance. 
In contrast, doubling the size of the baseline improves its cost but not its feasibility ($93.6\%$ vs. $94.2\%$).
Overall, the main result supports our central hypothesis: separating timescales 
improves both the closed-loop performance and the fidelity to $\pi_\MPC$.
This suggests that the main bottleneck for a more efficient BC
controller is not only the network's complexity, which is larger for the second baseline,
but also the quality and variety of the learned representations.

\begin{table}[t]
\centering
\renewcommand{\arraystretch}{1.25}
\resizebox{\linewidth}{!}{%
{\small
\begin{tabular}{|c|cc||cc|}
\specialrule{1.2pt}{0pt}{0pt}
\textbf{Method}
& \multicolumn{2}{c||}{\textbf{Standalone}}
& \multicolumn{2}{c|}{\textbf{Safety-Augmented~\cite{hose2025approximate}}}
\\
\specialrule{0.8pt}{0pt}{0pt}
\textbf{Embed. (Size)}
& \textbf{Cost Ratio} & \textbf{Const. Satisfaction}
& \textbf{Cost Ratio} & \textbf{Const. Satisfaction}
\\
\specialrule{1.2pt}{0pt}{0pt}
BC Baseline (64) & $4.71 \pm 1.47$ & $10.2 \pm 6.7\%$ & $1.260 \pm 0.092$ & $98.7 \pm 0.5\%$ \\
BC Baseline (128) & $4.30 \pm 1.20$ & $11.8 \pm 5.7\%$ & $1.216 \pm 0.055$ & $98.7 \pm 0.7\%$ \\
\specialrule{0.8pt}{0pt}{0pt}
\rowcolor{oursblue}
SRLBC (64) & $\mathbf{2.89 \pm 0.76}$ & $24.8 \pm 11.5\%$ & $\mathbf{1.138 \pm 0.039}$ & $98.6 \pm 0.8\%$ \\
\rowcolor{oursblue}
SRLBC (64) + Decor. & $3.18 \pm 0.95$ & $\mathbf{25.6 \pm 7.2\%}$ & $1.183 \pm 0.080$ & $98.5 \pm 0.7\%$ \\
\specialrule{0.8pt}{0pt}{0pt}
NMPC & $\mathit{1.00}$ & $\mathit{100.0\%}$ & $\mathit{1.00}$ & $\mathit{100.0\%}$ \\
\specialrule{1.2pt}{0pt}{0pt}
\end{tabular}
}}
\caption{Out-of-distribution evaluation on 175 feasible transients: cost ratio to the NMPC controller and constraint 
satisfaction, in standalone and under safety augmentation. $\mathrm{Mean} \pm \mathrm{Std}$ over 20 training seeds.}
\label{tab:ood}
\end{table}

\paragraph{Out-of-distribution (OOD) results}
We then evaluated the same policies, without retraining, on 175 feasible OOD transients (Table~\ref{tab:ood}). 
The performance of standalone policies heavily degrades under this shift: their closed-loop cost reaches 3 to 5 times that of the NMPC controller, 
and at most a quarter of the trajectories remain feasible. 
SRLBC still degrades markedly less than the baselines, with twice the feasibility and 
an excess cost almost halved (Holm-corrected $p < 10^{-2}$), but no learned policy can be deployed standalone in this regime. 

\paragraph{Safety-augmented results}
We implemented the safety-augmented pipeline from~\cite{hose2025approximate}. 
At each timestep, the input sequence predicted by the policy is simulated over the prediction horizon and applied if it satisfies $h \leq 0$ 
at every step. Otherwise, the NMPC controller is called. 
Since computing a terminal set is out of reach for this high-dimensional stiff system, 
we use it without terminal ingredients, i.e., without recursive feasibility guarantees.
In distribution (Table~\ref{tab:quantitative_results}, right), it drastically improves 
feasibility (from $94$--$97\%$ to $99.8$--$99.9\%$), while maintaining a $\sim 7\times$ acceleration factor across all policies
due to a small rejection rate ($\sim 1.6\%$). 
In this context, SRLBC still outperforms both baselines, with a lower cost ratio to the NMPC controller ($1.053$ vs. $1.074$ and $1.094$, Holm-corrected $p < 10^{-5}$).
Out of distribution, the rejection rate only doubles (from $1.6\%$ to $3.1\%$), 
feasibility returns to $98.5$--$98.7\%$ for all architectures (Table~\ref{tab:ood}, right), 
and the controller remains $4.8\times$ faster than the NMPC controller on average. 
Again, SRLBC retains the lowest cost ratio ($1.138$ vs. $1.216$ and $1.260$ for the baselines, Holm-corrected $p < 10^{-3}$). 
The decorrelated variant, however, loses most of this advantage out of distribution ($1.183$).

\paragraph{Computational time and hardware}
Roll-outs using the NMPC controller yield its trajectory 
in 5 to 6 minutes for 48h simulations. 
In comparison, roll-outs using any BC-based controller are rapid: the online evaluation 
of the controller is almost instantaneous, and does not depend on the exogenous input sequence. 
Hence, the roll-outs using standalone BC policies are always obtained in approximately 20 seconds, which is almost fully due to the 
simulation cost of the trajectories. 
This yields an acceleration of a factor of $\sim$15 compared to the usual NMPC 
controller (see Table~\ref{tab:quantitative_results}).
All closed-loop roll-outs, for both the NMPC controller and the learned controllers, were run on a single core 
of an Intel Xeon Platinum 8260 (2.40 GHz), so that computation times are directly comparable. 
These timings are measured on a workstation CPU, whereas the target is an embedded platform 
with much tighter memory and compute budgets, where online optimization is expected to be much slower: 
the relevant quantity is the number of optimizer calls avoided (up to $98.4\%$), rather than the absolute timings. 
Each network was trained with PyTorch~\cite{paszke2019pytorch} on a single NVIDIA V100 GPU, 
in less than 10 minutes per model.

\begin{figure}[!htb]
    \centering
    \begin{subfigure}{0.49\linewidth}
        \centering
        \includegraphics[width=\linewidth]{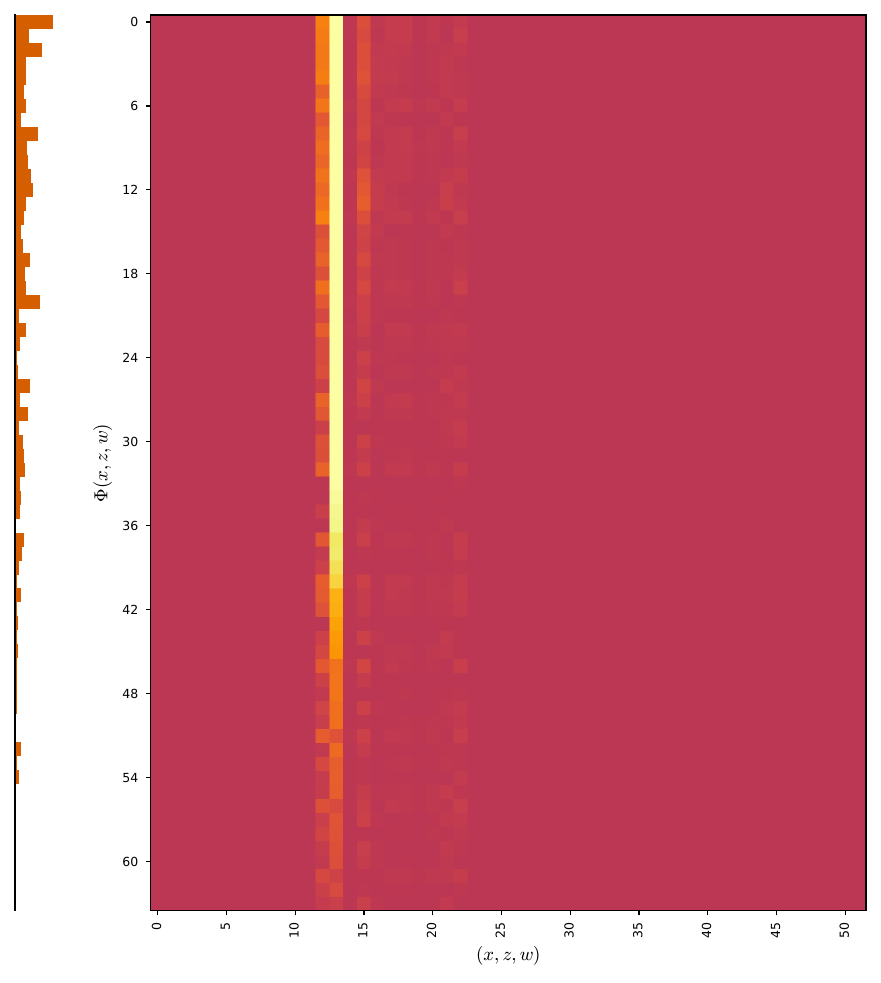}
        \caption{Shared: Jacobian.}
    \end{subfigure}
    \hfill
    \begin{subfigure}{0.49\linewidth}
        \centering
        \includegraphics[width=\linewidth]{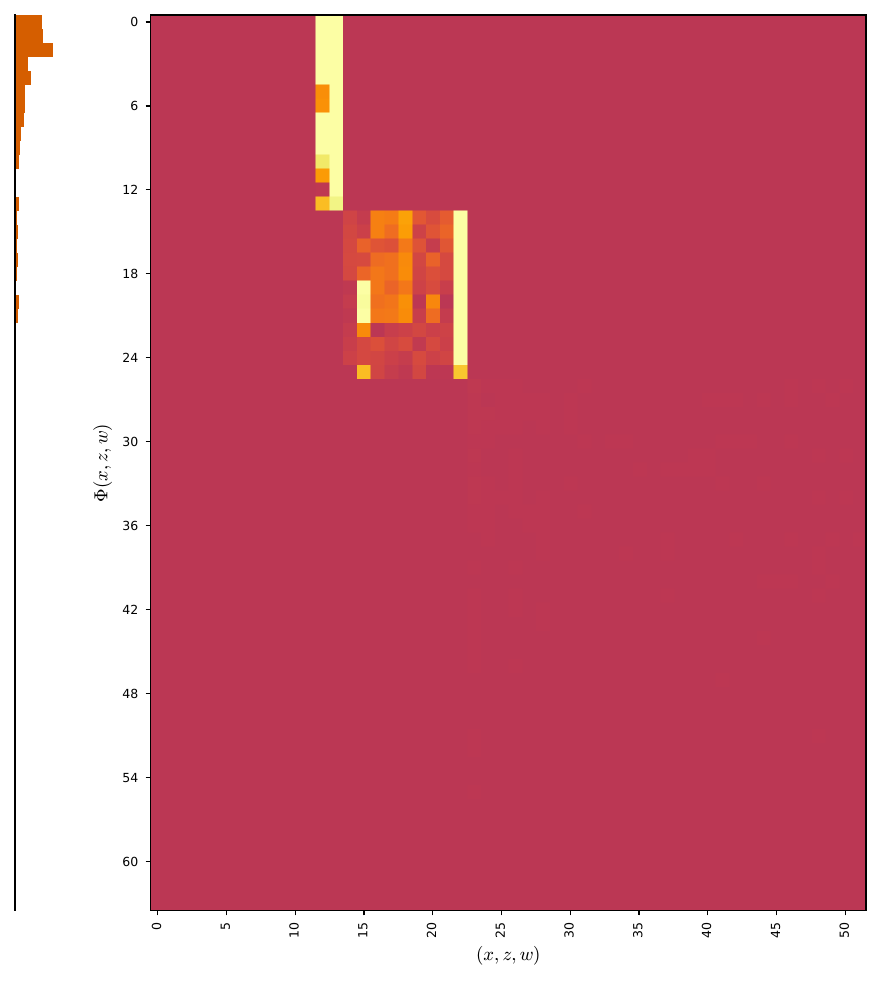}
        \caption{SRLBC + Decor.: Jacobian.}
    \end{subfigure}

    \vspace{0.3em}
    \begin{subfigure}{0.49\linewidth}
        \centering
        \includegraphics[width=\linewidth]{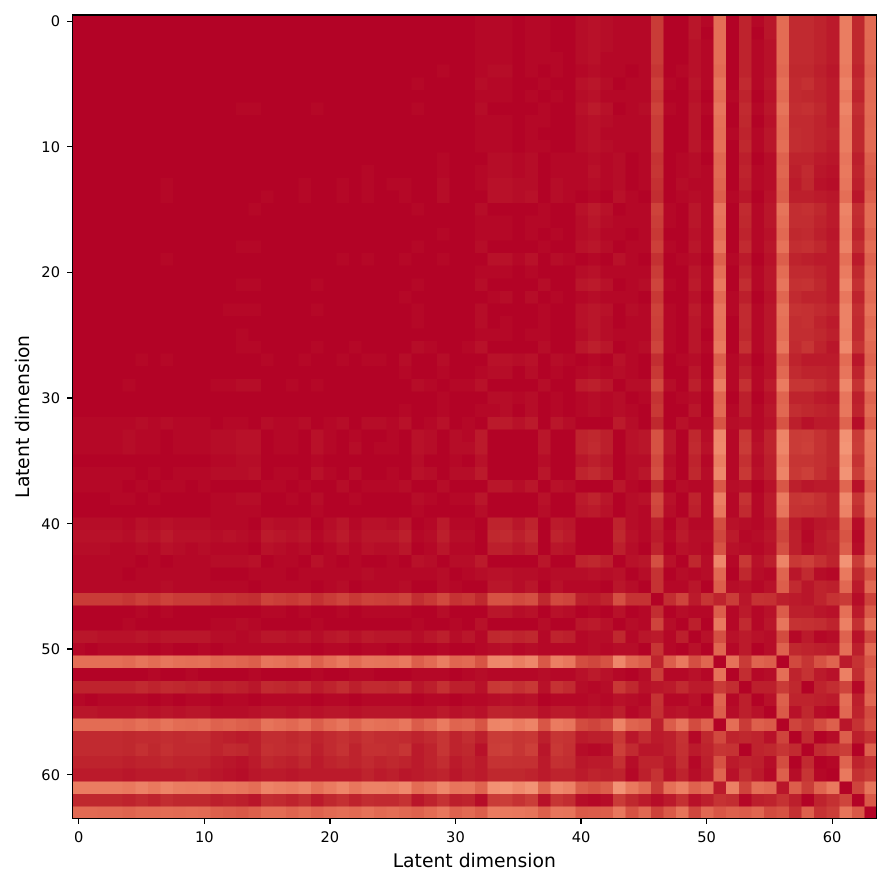}
        \caption{Shared: correlation.}
    \end{subfigure}
    \hfill
    \begin{subfigure}{0.49\linewidth}
        \centering
        \includegraphics[width=\linewidth]{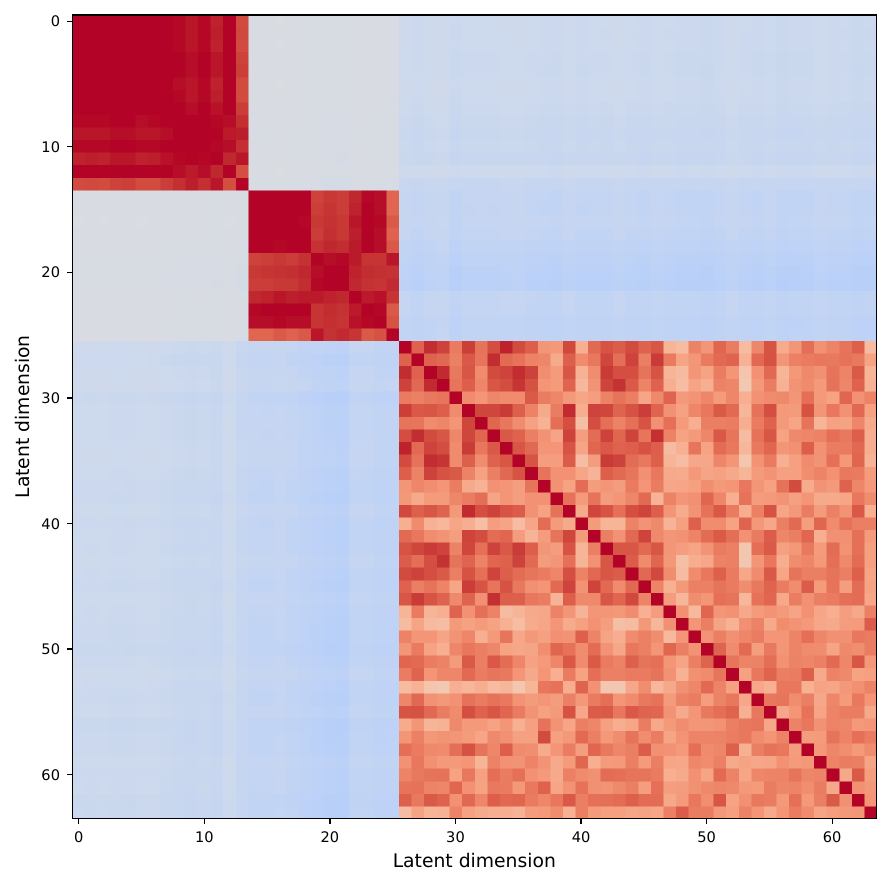}
        \caption{SRLBC + Decor.: correlation.}
    \end{subfigure}
    \caption{\textbf{Top:} averaged absolute Jacobian coefficients of each embedding (normalized
    by the input variances); left histograms give the relative importance of the $i^\mathrm{th}$ component 
    $[\Phi(x, z, w)]_i$ for the controller output $\pi_\BC \circ \Phi(x, z, w)$.
    \textbf{Bottom:} correlation matrices between the components of each embedding.}
    \label{fig:sensitivity_analysis}
\end{figure}

\paragraph{Analysis of the learned representations and interpretability of the controller}
With SRLBC, one can more easily see which physical quantities the controller relies on. 
With a shared embedding, this information is distributed across all latent features.
In Fig.~\ref{fig:sensitivity_analysis}, we plot the Jacobian matrices of the embeddings for both the shared and the 
separated encoders, the Jacobian of the controller with respect to these embeddings, 
and the correlation matrices between the latent components.
By design, the separated architecture imposes a block-structured decomposition of the latent space, 
since $\frac{\partial \phi_x}{\partial [z, w]} = 0_{d_x \times (n_z + n_w)}, \frac{\partial \phi_z}{\partial [x, w]} = 0_{d_z \times (n_x + n_w)}$, 
and $\frac{\partial \phi_w}{\partial [x, z]} = 0_{d_w \times (n_x + n_z)}$.
This structure is not a finding in itself, but it makes each latent component 
attributable to a single group of physical variables.
According to the Jacobian sensitivity analysis, the output of the controller is mainly driven 
by the contributions of the boron concentration (which is natural as the controller injects boron or water into the core) 
and the fast state variable (which is the only state argument of the cost and constraint functions).
The rest of the state and exogenous variables mostly help regularize.
More strikingly, $\frac{\partial \pi}{\partial \Phi}$ is much more concentrated for SRLBC, 
whereas the shared embedding spreads the contributions over many components.
Its correlation matrix also reveals a much more redundant representation, 
while enforcing decorrelation reduces the correlations within each block.

\begin{table}[t]
\centering
\renewcommand{\arraystretch}{1.25}
\resizebox{\linewidth}{!}{%
{\small
\begin{tabular}{|c|cc||cc|}
\specialrule{1.2pt}{0pt}{0pt}
\textbf{Method}
& \multicolumn{2}{c||}{\textbf{In distribution}}
& \multicolumn{2}{c|}{\textbf{Out of distribution}}
\\
\specialrule{0.8pt}{0pt}{0pt}
\textbf{Embed. (Size)}
& \textbf{Standalone} & \textbf{Safety-Aug.~\cite{hose2025approximate}}
& \textbf{Standalone} & \textbf{Safety-Aug.~\cite{hose2025approximate}}
\\
\specialrule{1.2pt}{0pt}{0pt}
BC Baseline (64) & $1.44 \pm 0.14$ & $0.994 \pm 0.003$ & $24.1 \pm 12.2$ & $1.001 \pm 0.005$ \\
BC Baseline (128) & $1.34 \pm 0.13$ & $0.995 \pm 0.003$ & $18.0 \pm 9.3$ & $1.000 \pm 0.004$ \\
\specialrule{0.8pt}{0pt}{0pt}
\rowcolor{oursblue}
SRLBC (64) & $\mathbf{1.20 \pm 0.06}$ & $0.995 \pm 0.004$ & $12.0 \pm 7.6$ & $1.003 \pm 0.003$ \\
\rowcolor{oursblue}
SRLBC (64) + Decor. & $1.21 \pm 0.08$ & $0.996 \pm 0.001$ & $\mathbf{10.2 \pm 7.0}$ & $1.003 \pm 0.005$ \\
\specialrule{0.8pt}{0pt}{0pt}
NMPC & $\mathit{1.00}$ & $\mathit{1.00}$ & $\mathit{1.00}$ & $\mathit{1.00}$ \\
\specialrule{1.2pt}{0pt}{0pt}
\end{tabular}
}}
\caption{Constraint violation ratio $\|h_+(\pi_\BC \circ \Phi)\| / \|h_+(\pi_\MPC)\|$ relative to that of the NMPC controller, on transients 
that are infeasible even for the latter. $\mathrm{Mean} \pm \mathrm{Std}$ over 20 training seeds.}
\label{tab:infeasible}
\end{table}

\paragraph{Behavior on infeasible transients}
In operation, a power demand may exceed what the reactor can follow within the safety margins. 
We therefore also compared the policies on a separate set of 500 aggressive transients 
that are infeasible even for the NMPC controller. 
Since no controller is feasible there with respect to the strict safety constraints, we compare the relative constraint violation 
$\|h_+\|$ to that of the NMPC controller (Table~\ref{tab:infeasible}). 
In standalone, SRLBC exceeds this violation by about $20\%$, against $34$ to $44\%$ for the baselines. 
Under the safety-augmented pipeline, all policies match the NMPC controller: when the demand leaves the admissible domain, 
the certified controllers do not degrade safety further than the NMPC itself. 
The cost ordering is also preserved (not shown in Table~\ref{tab:infeasible}: cost ratio of $1.05$ for SRLBC vs. $1.11$ for the baselines in standalone, 
and $1.012$ vs. $1.017$--$1.025$ under safety augmentation). 
The same holds on infeasible out-of-distribution transients: standalone violations reach 10 to 24 times that of the NMPC controller 
(SRLBC variants reducing that of the baselines by a third to a half) while safety augmentation again matches it.

\section{Discussion}\label{sec:Discussion}


\paragraph{What safety augmentation does and does not equalize}
Once a certification step is added, feasibility, computation time and rejection rate become nearly independent of the architecture, 
in distribution as well as out of distribution. 
Although SRLBC imitates the NMPC controller more accurately, all policies are rejected at almost the same rate, 
which suggests that rejections are driven by a few intrinsically hard situations rather than by the average imitation error. 
However, what the safety filter does not equalize is closed-loop performance: it only checks the safety of the proposed input sequence, but its quality
remains closely related to the policy. 
The benefit of a structured representation therefore remains visible in the closed-loop cost, 
and even more so for transients outside the training distribution. 
Overall, the filter provides safety, while the policy provides performance.

\paragraph{Limitations and perspectives}
First, without terminal ingredients, the safety-augmented pipeline provides no recursive feasibility guarantee: 
in fact, feasibility remains slightly below that of the NMPC controller, and the certification step relies on the simulation model being accurate. 
Second, the decorrelation penalty improves the structure of the representation but has very limited impact on the closed-loop performance.
It is even less robust out of distribution. 
Finally, the timescale partition must be known a priori, and the latent dimensions of the three encoders have to be selected by expert knowledge 
or prior hyperparameter optimization (as we did). 
Still, designing simpler and more interpretable BC controllers is a further step towards 
autonomous explicit feedback laws of the form $u = \pi(x)$ approximating optimization-based controllers
as is traditional in control theory.
Generalization to systems with three or more timescales could be done in the future by
similarly designing a timescale-aware encoder, allowing further interpretability
and more efficient end-to-end controllers.
Statistical learning bounds on the deviation between SRLBC and the NMPC controller could also bound the rejection rate, 
and hence the computational load of the fall-back phase, enabling provably safe and fast controllers. 
Industrially, this is particularly relevant to deploy new online control strategies (e.g., with different cost or constraint functions) 
that currently remain too computationally expensive.

\section{Conclusion}

In this paper, we introduced SRLBC, a simple architecture for designing 
behavioral clones of experts for multiscale nonlinear control systems. 
Our results suggest that, for a network expressive enough to approximate an NMPC controller, 
further gains can come from structuring the controller itself.
Enforcing physics-based prior knowledge through a structured embedding layer for end-to-end BC
enhances performance and interpretability of the resulting controller on a challenging PWR-simulator benchmark.
Our methodology achieves lower integral cost, constraint violations, and more accurate imitation of the 
expert while maintaining a structured embedding, consistently across 20 training seeds. 
Combined with the safety-augmented approximate NMPC framework of~\cite{hose2025approximate}, 
the learned policies run up to 7 times faster than the NMPC controller with a feasibility of at least $99.8\%$, 
SRLBC retaining the lowest closed-loop cost.
Out of distribution, standalone policies fail, but the certified combination remains feasible on more than $98\%$ of the trajectories, 
again with the lowest cost for SRLBC.



\bibliographystyle{IEEEtran}
\bibliography{refs.bib}               
                                                   






\end{document}